\documentclass[10pt,twocolumn,letterpaper]{article}

\usepackage{wacv}
\usepackage{times}
\usepackage{epsfig}
\usepackage{graphicx}
\usepackage{amsmath}
\usepackage{amssymb}
\usepackage{booktabs}
\usepackage{dcolumn}
\newcommand{\R}{\mathbb{R}}
\def\wacvPaperID{5} % Enter the WACV Paper ID here

\wacvfinalcopy % *** Uncomment this line for the final submission

\ifwacvfinal
\fi

\ifwacvfinal
\usepackage{hyperref}
\hypersetup{breaklinks=true,bookmarks=false}

\else
\usepackage{hyperref}
\hypersetup{pagebackref=true,breaklinks=true,colorlinks,bookmarks=false}

\fi

\ifwacvfinal
\else
\fi

\begin{document}

%%%%%%%%% TITLE
\title{Pose-based Sign Language Recognition using GCN and BERT }

\author{Anirudh Tunga\thanks{equal contribution}\\
Purdue University\\
{\tt\small atunga@purdue.edu}
\and
Sai Vidyaranya Nuthalapati*\\
{\tt\small vidyaranya.ns@gmail.com}

\and
Juan Wachs\\
Purdue University\\
{\tt\small jpwachs@purdue.edu}
}

\maketitle
%\thispagestyle{empty}

%%%%%%%%% ABSTRACT
\begin{abstract}
   Sign language recognition (SLR) plays a crucial role in bridging the communication gap between the hearing and vocally impaired  community and the rest of the society. Word-level sign language recognition (WSLR) is the first important step towards understanding and interpreting sign language. However, recognizing signs from videos is a  challenging task as the meaning of a word depends on a combination of subtle body motions, hand configurations and other movements. Recent pose-based architectures for WSLR either model both the spatial and temporal dependencies among the poses in different frames  simultaneously or only model the temporal information without fully utilizing the spatial information.

   We tackle the problem of WSLR using a novel pose-based approach, which captures spatial and temporal information separately and performs late fusion. Our proposed architecture explicitly captures the spatial interactions in the video using a Graph Convolutional Network (GCN). The temporal dependencies between the frames are captured using Bidirectional Encoder Representations from Transformers (BERT). Experimental results on WLASL, a standard word-level sign language recognition dataset show that our model significantly outperforms the state-of-the-art on pose-based methods by achieving an improvement in the prediction accuracy by up to 5\%.
\end{abstract}

%%%%%%%%% BODY TEXT
\section{Introduction}

Hearing and vocally impaired people use sign language instead of spoken language for communication. Just like any other language, sign language has an underlying structure, inter alia, grammar and intricacies to allow users (signers or interpreters) to fully express themselves. To comprehend sign language, one must consider and understand multiple aspects such as hand movements, shape and orientation of the hand, shoulder orientation, head movements and facial expressions. The study of accurately recognizing and understanding sign language technique falls under the ambit of sign language recognition.

According to \cite{how_manyu_sign}, there are approximately 500,000 users of American Sign Language in the US itself. While on one hand, hearing and vocally impaired communities are completely dependent on sign language for communication, on the other hand, the rest of the world does not understand sign language, creating a communication barrier between the two groups. It is also unlikely that people without such impairments will learn an additional language which is not seen as a necessity for them. This gap between the rest of the world and the hearing and vocally impaired community can be reduced by developing Automatic Sign Language Recognition (ASLR). 
\begin{figure}[t]
\begin{center}
   \includegraphics[width=1\linewidth]{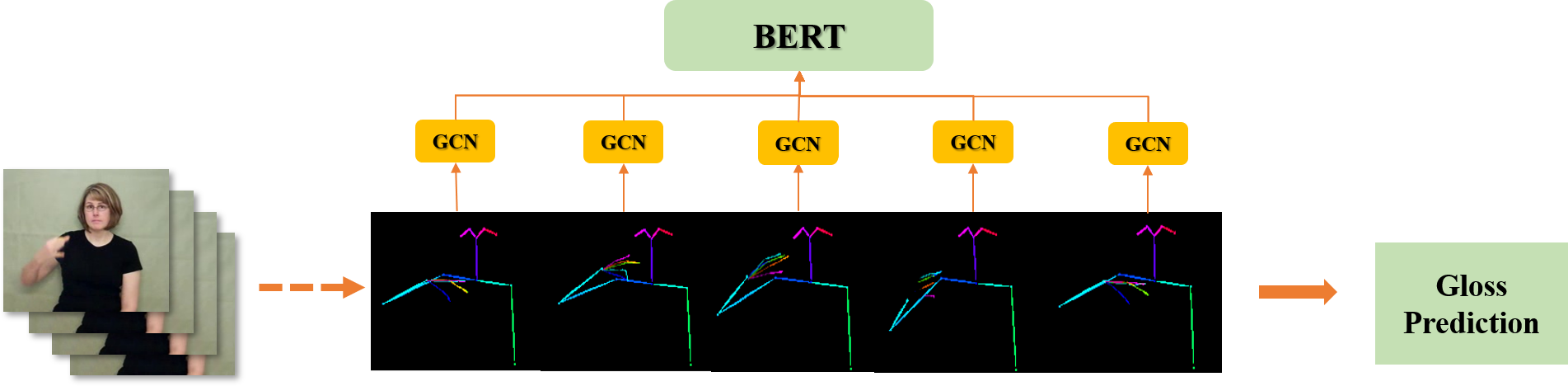}
\end{center}
   \caption{We train a model based on GCN and BERT to predict the glosses from the poses extracted from video frames.}
\label{fig:intro}
\vspace{-15pt}
\end{figure}

Sign language recognition can be broadly classified into two parts: word-level sign language recognition (WSLR) and sentence-level sign language recognition. WSLR is the fundamental building block for interpreting sign language sentences. As shown in Figure \ref{fig:intro}, signalling a sign language word requires very subtle body movements that makes WSLR a particularly challenging problem. In this paper, we focus on WSLR by exploiting the information from human skeletal motion. In WSLR, given a sign language video, the goal is to predict the word that is being signalled in the video. `Gloss' is another term for representing the word that is being shown. Recently, deep learning techniques have shown a huge promise in the field of WSLR \cite{pigou, msasl, wlasl, kishore}. The techniques that are employed for ASLR can be divided into two categories: 1. Methods based on 2D pose estimation, and 2. Architectures utilizing the holistic image features. We believe that human skeletal motion plays a significant role in conveying what word the person is signalling. Hence, this work focuses on a pose-based model to tackle the problem of WSLR. The existing pose-based methods either model both the spatial and temporal dependencies between the poses in different frames simultaneously or only model the temporal information without fully utilizing the spatial information  \cite{wlasl}. Inspired by \cite{kalfaoglu2020late}, where the authors have used late temporal fusion to achieve a performance boost in action recognition, we propose a novel pose-based architecture, GCN-BERT, which first captures the spatial interactions in every frame comprehensively before explicitly utilizing the temporal dependencies between various frames in the video. We validate our architecture on the recently released large-scale WLASL dataset \cite{wlasl} and experimental results show that the proposed model significantly outperforms the state-of-the-art pose-based models by achieving an accuracy improvement of up to 5\%.
%-------------------------------------------------------------------------

\section{Related Work}
Sign language recognition mainly involves three phases - feature extraction phase, temporal modelling phase, and prediction phase. Historically, spatial representation was generated using hand crafted features like HOG-based features \cite{buehler2009learning,cooper2012sign}, SIFT-based features \cite{yang2010chinese,tharwat2015sift}, and frequency domain features \cite{al2009video,badhe2015indian}. Temporal modelling was done using Hidden Markov Models (HMM) \cite{starner1998real,evangelidis2014continuous,zhang2016chinese}, and hidden conditional random fields \cite{wang2006hidden}. Some works utilized Dynamic Time Wrapping \cite{sakoe1978dynamic,lichtenauer2008sign} to handle varying frame rates. The prediction phase was treated as a classification problem, and models like Support Vector Machine (SVM) \cite{nagarajan2013static} were used to predict the words from the signs. The vast majority of traditional sign language recognition models are evaluated on small scale datasets, which had less than one hundred words \cite{zafrulla2011american,lim2016block,kulkarni2010appearance}.

With the advent of deep neural networks, there was a significant boost in the performance for many video-based tasks like action recognition \cite{donahue2015long,feichtenhofer2016convolutional}, and gesture recognition \cite{camgoz2016using}. Both, action recognition and sign language recognition share a similar problem structure. Inspired from the network architectures for action recognition, new architectures for sign language recognition were proposed. For example, a CNN-based architecture was used for sign language recognition in \cite{pigou2014sign}, and a  frame-based CNN-HMM model for sign language recognition was proposed in \cite{koller2016deep}. These two papers are representative of a more general trend of deep neural-based architectures for sign language recognition. It was learnt that these works can be partitioned into two categories: image appearance based methods, and pose-based methods, which are presented in more detail below.

\subsection{Image appearance based methods}
Word level sign language recognition focuses mainly on intricate hand and arm movements, while the background is not very useful in recognition. In this section, we discuss some relevant image based methods for action recognition and sign language recognition.  

Utilizing the feature extraction capability of deep neural networks, Simonyan et al., \cite{simonyan2014two}, uses a 2D CNN to create a holistic representation of each input frame of the video and then uses those representations for recognition. Temporal dynamics of a video can be modelled by sequence modelling using recurrent neural networks.  The works \cite{yue2015beyond,donahue2015long}, use  Long  Short-Term Memory (LSTMs) to model the temporal dynamics of the features extracted through CNNs. In \cite{cui2017recurrent}, a  2D CNN-LSTM  architecture,  where, in  parallel  with  the  LSTMs, it also uses a  weakly  supervised  gloss-detection  regularization  network, consisting  of  stacked  temporal  1D  convolutions. A simpler variant of LSTMs, Gated-recurrent Units (GRU) \cite{cho2014properties}, which consist of only two gates  (update and reset gates), and have the internal state (output state) fully exposed, have also been used for temporal modelling \cite{yao2018action}. 

While the above works used RNNs to model the gesture temporal behaviour, a few works have used CNNs to achieve this. For instance, 3D CNNs \cite{taylor2010convolutional} can not only learn the holistic representation of each input frame, but also the spatio-temporal features. The  C3D \cite{tran2015learning} model was the first model to use 3D CNNs for action recognition. In \cite{joze2018ms}, the I3D \cite{carreira2017quo} architecture has been trained and adopted for sign language recognition. In \cite{zhou2019dynamic}, the authors extended the I3D architecture by adding a RNN. Recent work \cite{kalfaoglu2020late}, has used the Bidirectional Encoder Representations from Transformers (BERT) \cite{devlin2018bert} at the end of a 3D CNN.

\subsection{Pose-based methods}
\subsubsection{Pose estimation}
Human pose estimation involves localizing keypoints of human joints from a single image or a video. Historically, pictorial structures \cite{pishchulin2013poselet} and probabilistic graphical models \cite{yang2011articulated} were used to estimate the human pose. Recent advances in deep learning have greatly boosted the performance of human pose estimation. Two main methods exist in localizing the keypoints: directly regressing the $x,y$ coordinates of joints, and estimating keypoint heatmaps followed by a non-maximal suppression technique. In \cite{toshev2014deeppose}, Toshev et al. introduced `Deep Pose', where they directly regress the keypoints from the frame. In \cite{tompson2014joint}, Tompson et al. used  a ConvNet and a graphical model to estimate the keypoint heatmaps. Recent works \cite{newell2016stacked,cao2018openpose} have improved the performance of human pose estimation significantly using heatmap estimation. Though pose estimation succeeds at estimating positions of human joints, it does not explore the spatial dependencies among these estimated keypoints or joints.

\subsubsection{Adapting pose estimation for activity recognition}
Human poses and their dependencies can be used to recognize actions. In \cite{Nie_2015_CVPR}, the authors built a graphical model using poses to recognize actions. In \cite{Cheron_2015_ICCV}, the authors used pose-based feature descriptors for activity recognition. RNNs have been used to model the temporal sequential information of the  pose movements, and the representation output by RNN has been used for the sign recognition. In \cite{Martinez_2017_CVPR}, the authors analyze human motions by using RNNs to model the pose sequences. A few works, for instance, \cite{Li_2019_CVPR}, have used graph network based architectures to model the spatial and temporal dependencies of the pose sequence. 

Recently, pose-based methods have been used for sign language recognition. In \cite{wlasl}, the authors used the pose sequence extracted from the video followed by a graph convolution network to model the spatio-temporal dependency for sign language recognition. However, the existing works on sign language recognition, either model the spatial and temporal dependencies together or ignore the spatial dependencies, and only model the temporal dependencies. In order to overcome these limitations, we propose a model where we extract the spatial and temporal encoding separately and comprehensively use these two different components in our architecture.

\section{Proposed Architecture}

\subsection{Notation}
We denote the WSLR dataset containing \textit{N} labelled training examples by $\mathcal{D}$ = $\{X_i, Y_i\}_{i=1}^{N}$ where $X_i \in \R^{l*w*h*3}$ is the input RGB video; \textit{l, w, h} represent the length, width and height of the video respectively; and $Y_i \in [1,2,...,G]$ is the sign corresponding to the input video, and \textit{G} is the number of output classes.

\subsection{Overview}
Sign languages are based on a sequence of body part movements to convey a message. Deep learning methods that are trained on skeletal information have shown a great promise in detecting and analyzing such body movements \cite{kindir}. This is primarily because using poses helps the model to focus on the most important parts of the image rather than focusing on the unimportant components (for e.g. lighting conditions, and background of the image).  Pose-based approaches have been widely used in the literature to extract such skeletal information. They mainly use RNNs to capture the temporal information of the changes in the pose. However, such pose-based approaches fail to capture the spatial interactions between various keypoints in the pose which is extremely important for sign language recognition.  Deviating from the existing pose-based architectures which model the temporal and spatial interactions together, we delegate the modelling of the two interactions to two individual components in our model. Inspired by the success of graph neural network in capturing the spatial information for sign language recognition, we use graph convolutional networks (GCN) \cite{wlasl, cao, kipf} to encode the spatial relationship among various body key points. Once the spatial information is collected across all the frames, we deploy BERT \cite{devlin2018bert}, a transformer-based architecture \cite{vaswani} to collate the temporal information. Figure \ref{fig:arch} depicts the GCN-BERT architecture in detail.

We represent a pose using \textit{K} keypoints representing the upper body and both the hands. Each keypoint is a 2-dimensional vector showing the location of the corresponding keypoint in the frame. These keypoints serve as inputs to our model. Further, as videos vary in length, we randomly select 50 consecutive frames from the input video to maintain a constant input size. This information is passed as input to our model for predicting the gloss.

\subsubsection{Pose-Based Graph Convolution Network}
We extract the spatial information from every video frame using a graph neural network. Inspired by  \cite{wlasl}, we use GCN to extract this information. The input to the graph is represented using $x_t \in X$ and $t \in [1,2,....,T]$, where \textit{T} is the number of frames in the video and $x_t \in \R^{K \times 2}$ (we multiply it with 2 as we use 2D coordinates for locating keypoints). We represent the human body as a fully connected graph \cite{chiu} which allow us to express the relative positions of various body keypoints, as essential parts in order to accurately determine  the gloss.  While the keypoints form the nodes in the graph, the edges are weighted and learnt during the training process. We represent the weigthed adjacency matrix as $A \in \R^{K \times K}$. Initially, the nodes are represented using the 2D coordinates for the corresponding keypoint. During the training process, the node representations are updated using the update rules below:
\begin{equation}
H^{(l + 1)} = f(H^{(l)}, A),
\end{equation}
\begin{equation}
f(H^{(l)}, A) = \sigma \left(AH^{(l)}W^{(l)} \right)
\end{equation}

where (i) $H^{(l)} \in \R^{K \times F}$ stands for the representation of nodes in \textit{l}-th neural network layer,  (ii) \textit{F} is the output feature from the previous layer.  Initially, \textit{H} is set to \textit{X} and so we take \textit{F} to be equal to 2,  (iii) $W^{(l)} \in \R^{(F \times F')}$ stands for the  weight matrix in \textit{l}-th neural network layer, and (iv) $\sigma$ represents a non-linear activation function which is  \textit{tanh} in our case. We can stack \textit{L} such layers to create accurate representations of each node in the graph i.e. $l \in [1,2,...,L]$. \textit{L} such stacked graph convolutional layers constitute one GCN network. The updated node representations encode spatial information of all the keypoints and interactions between them.
\begin{figure*}
\begin{center}
\includegraphics[scale=0.25]{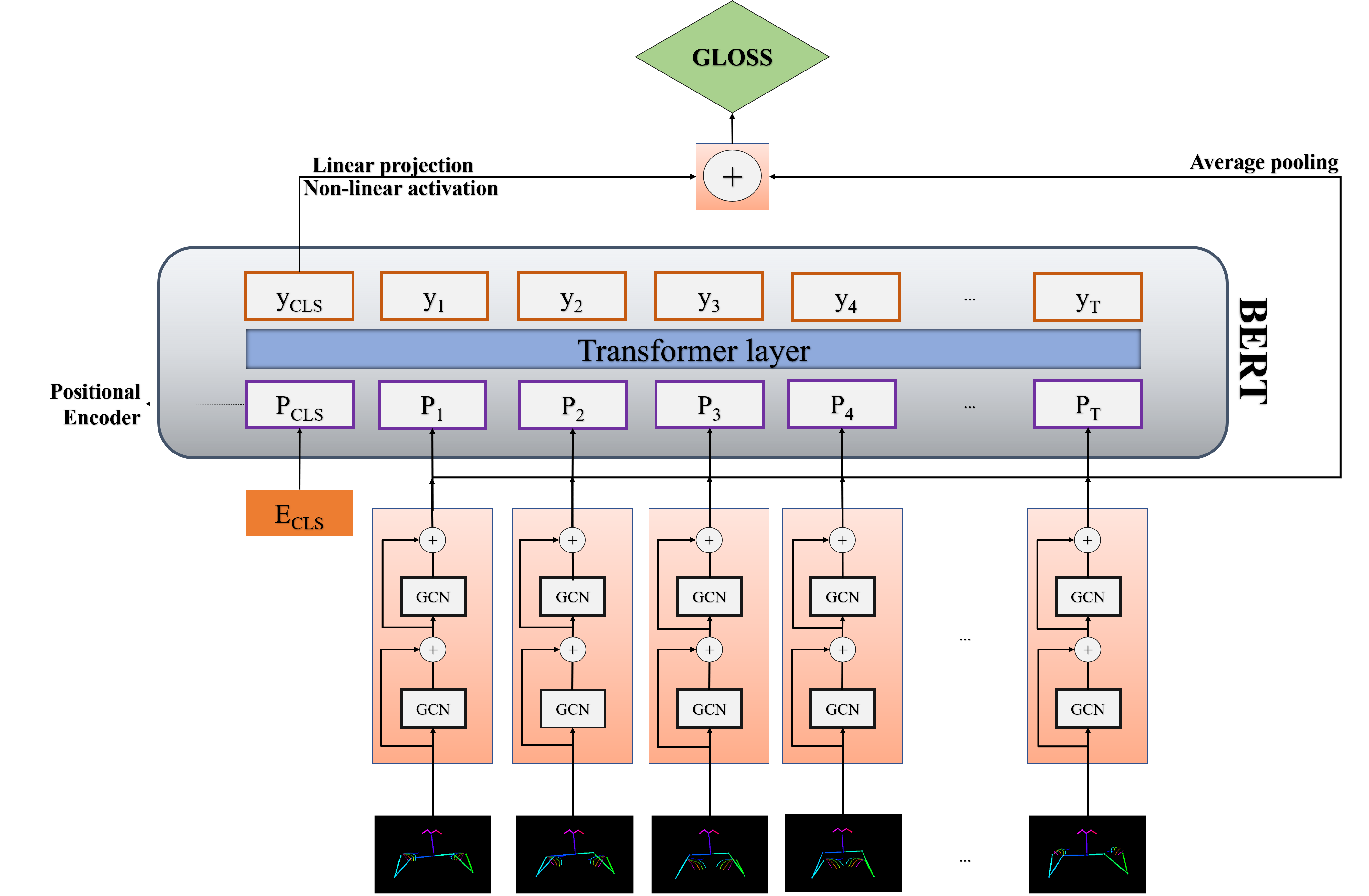}

\end{center}
   \caption{Illustration of the proposed GCN-BERT architecture. The poses extracted from the video are fed to the GCN to model spatial dependencies in the frames. This is followed by BERT to model the temporal dependencies between various frames in the video.}
\label{fig:arch}
\end{figure*}

Similar to \cite{wlasl}, we stack multiple GCN networks on top of each other and provide residual connections between the stacked GCNs. We represent the input to a single GCN network as \textit{I} and the output as $\tilde O$. With residual connections, the actual output of a single GCN network is given as follows:
\begin{equation}
    O = \tilde O + I
\end{equation}

This allows the network to learn to bypass a GCN network if required. We stack \textit{B} such networks to get the final output representations of keypoints. In Figure \ref{fig:arch}, we show the case where \textit{B} equals 2. While the process discussed above is for a single frame in the video, the same process can be repeated for all the frames in the video. In the end, we have the encoded spatial information for all the frames in the video denoted by $S \in \R^{T \times K \times F}$ where  $S = [H_1, H_2, \ldots, H_T] $. We also calculate the mean of all the spatial encodings along the temporal direction, denoted by $\hat{S} \in \R^{K \times F}$. This is followed by a fully connected layer and a non-linear activation to project into a \textit{G}-dimensional space, where \textit{G}, is the number of output classes. Let us denote the resultant encoding by $\hat{U}$. This will later be used to provide a skip connection from the output of GCN to the output of BERT.

\subsubsection{Temporal modelling using BERT}

Recently, architectures based solely on multi-head self-attention have achieved state-of-the-art results on sequence modelling tasks \cite{vaswani, lin, lin_wiki}. One such architecture - Bidirectional Encoder Representations from Transformers (BERT) \cite{devlin2018bert} - has shown a dramatic success in many downstream Natural Language Processing tasks. It has been designed to learn bidirectional representations by considering both the left and right contexts in all its layers. While it was initially introduced for NLP tasks, it is recently being used to model other sequential tasks such as action classification and video captioning \cite{sun_vidbert}.
\begin{table*}[ht!]
\caption{Top-1, top-5, top-10 accuracy (\%) achieved by pose-based models on WLASL dataset.}
\centering
\begin{tabular}{lc cccccc}
\toprule
& \multicolumn{3}{c}{WLASL100} & \multicolumn{3}{c}{WLASL300} \\
\cmidrule(r){2-4}\cmidrule(l){5-7}
Methods & {Top-1} & {Top-5}  & {Top-10} & {Top-1} & {Top-5}  & {Top-10} \\
\midrule
Pose-GRU \cite{wlasl} & 46.51 & 76.74 & 85.66 & 33.68 & 64.37 & 76.05 \\
Pose-TGCN \cite{wlasl}& 55.43 & 78.68 & 87.60 & 38.32 & 67.51 & 79.64 \\
GCN-BERT(ours) & \textbf{60.15} &\textbf{ 83.98} &\textbf{ 88.67} & \textbf{42.18} &\textbf{ 71.71 }&\textbf{ 80.93}\\
\bottomrule
\end{tabular}
\label{tab:table_pose_results}
\end{table*}
Inspired by the success of BERT in problems related to activity recognition \cite{kalfaoglu2020late}, we use BERT to learn bidirectional representations over sequence of encoded spatial information $S$ generated from GCN. This enables the model to learn contextual information from both left and right  directions. Similar to \cite{devlin2018bert}, the input $S$ is concatenated with learned position embeddings (denoted by $P_i$ for \textit{i}-th input position.) to capture the positional information. Then, we add a classification token $s_{cls}$ to the start of the input. The corresponding output from the last layer in BERT, $y_{cls}$ is passed through a fully connected layer and is eventually used for predicting the gloss.

Single head self-attention in a BERT layer computes the output as follows \cite{devlin2018bert, kalfaoglu2020late}:

\begin{equation}
    M\left(s_i\right) = \left( \frac{1}{N(s)}  \sum_{\forall j} V(s_j)f(s_i, s_j)  \right)
    \label{eq:first}
\end{equation}
where $s_i \in S$ represents the spatial information corresponding to \textit{i}-th pose extracted from GCN. $N(s)$ is the normalization factor and is used to produce a softer attention distribution and to avoid extremely small gradients \cite{vaswani}. ${f(s_i, s_j)}$ is used to measure the similarity between $s_i, s_j$ and is defined as $softmax_j(Q(s_i)^{T}K(s_j)),$ where the functions \textit{Q} and \textit{K} are learned linear projections. Combined with \textit{V}, which is also a learned linear projection, the functions \textit{Q} and \textit{K} project the inputs to a common space before applying the similarity measure. 

The single head self-attention sub-layer computation above predominantly consists of linear projects. To add non-linearity to the model, we use Position-wise Feed-Forward Network (PFFN) to the outputs of the self-attention sub-layer identically and separately at each position.
\begin{equation}
\begin{gathered}
\begin{split}
    PFFN(x) = \mathbf {W_2}  \textit{GELU}(\mathbf{W_1}x + \mathbf{b_1}) + \mathbf{b_2}
    \\
GELU(x) = x\phi(x)
\end{split}
\end{gathered}
\label{eq:pffn}
\end{equation}

where $\phi(x)$ represents the cumulative distribution function of the standard Gaussian distribution and $\mathbf{W_1}, \mathbf {W_2}, \mathbf{b_1}, \mathbf{b_2}$ are learnable parameters. Combining Equations \ref{eq:first} and \ref{eq:pffn}, we calculate $y_i$ as follows:
\begin{equation}
y_{i} = PFFN(M(x_{i})).
\label{eq:ycls}
\end{equation}
While the Equations \ref{eq:first}, \ref{eq:pffn}, \ref{eq:ycls} show attention calculation for single head, we can calculate attention using multiple heads and average the outputs. This constitutes a transformer layer.

Using the equations above we calculate $y_{cls}$, the output from the transformer layer corresponding to $x_{cls}$, which is passed through a fully connected layer projecting it into a \textit{G}-dimensional space followed by \textit{tanh} activation . Let us denote the resulting spatial-temporal encoding by $\hat{V}$. 

We provide a skip connection from the output of GCN to the output of BERT as follows:
\begin{equation}
    \hat{y} = \hat{U} + \hat{V},
\end{equation}
which is followed by a softmax layer to predict the output label. We use the standard cross-entropy loss to train the neural network.

\section{Experiments and Analysis}

In this section we describe the experimental setup, and provide quantitative and qualitative results. 

\subsection{Dataset}
\begin{table}[h!]
  \begin{center}
    \caption{Dataset statistics}
    \label{tab:table1}
    \begin{tabular}{l|c|c|c|r} % <-- Alignments: 1st column left, 2nd middle and 3rd right, with vertical lines in between
      \textbf{} & {Classes} & {Train} & {Validation} & {Test} \\
      \hline
      WLASL100 & 100 & 1442 & 338 & 258\\
      WLASL300 & 300 & 3548 & 901 & 668\\
    \end{tabular}
  \end{center}
  \label{tab:ds_stats}
\end{table}

The dataset used to validate the results in the paper is the Word Level American Sign Language (WLASL) dataset \cite{wlasl}. This dataset has been recently introduced and supports large scale WSLR. The videos contain native American Sign Language (ASL) signers or interpreters, showing signs of a specific English word in ASL. We show the dataset split of WLASL in Table \ref{tab:table1} \cite{wlasl_cvpr}.The number of classes represents the number of different glosses in the dataset.  For our experiments, we use the public dataset split released by the dataset authors.

\begin{figure*}
\begin{center}
\includegraphics[scale=0.8]{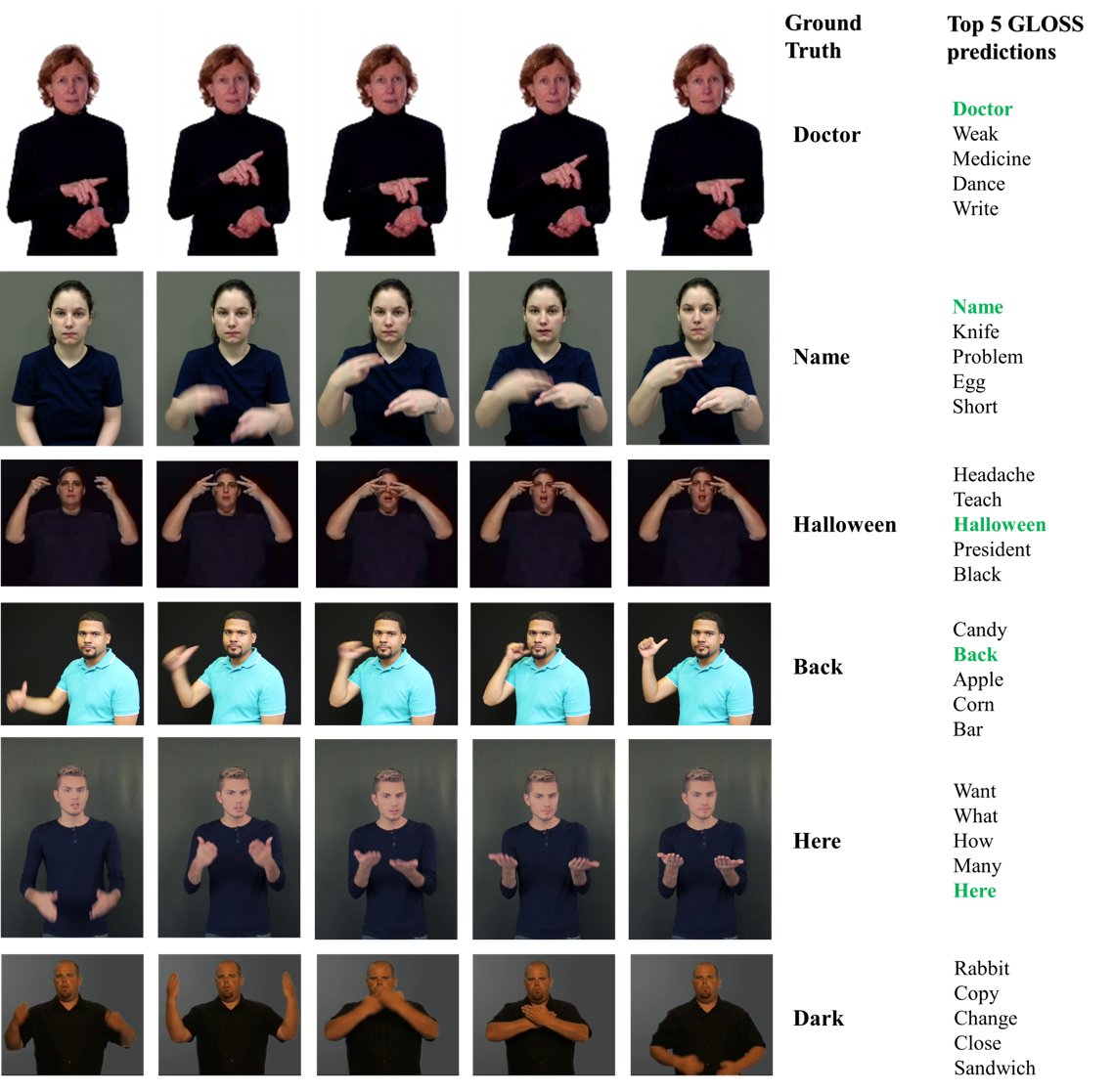}

\end{center}
   \caption{Left to right: video frames, ground truth, and predicted glosses for various videos.}
\label{fig:res}
\end{figure*}

\begin{figure*}
\begin{center}
\includegraphics[scale=0.55]{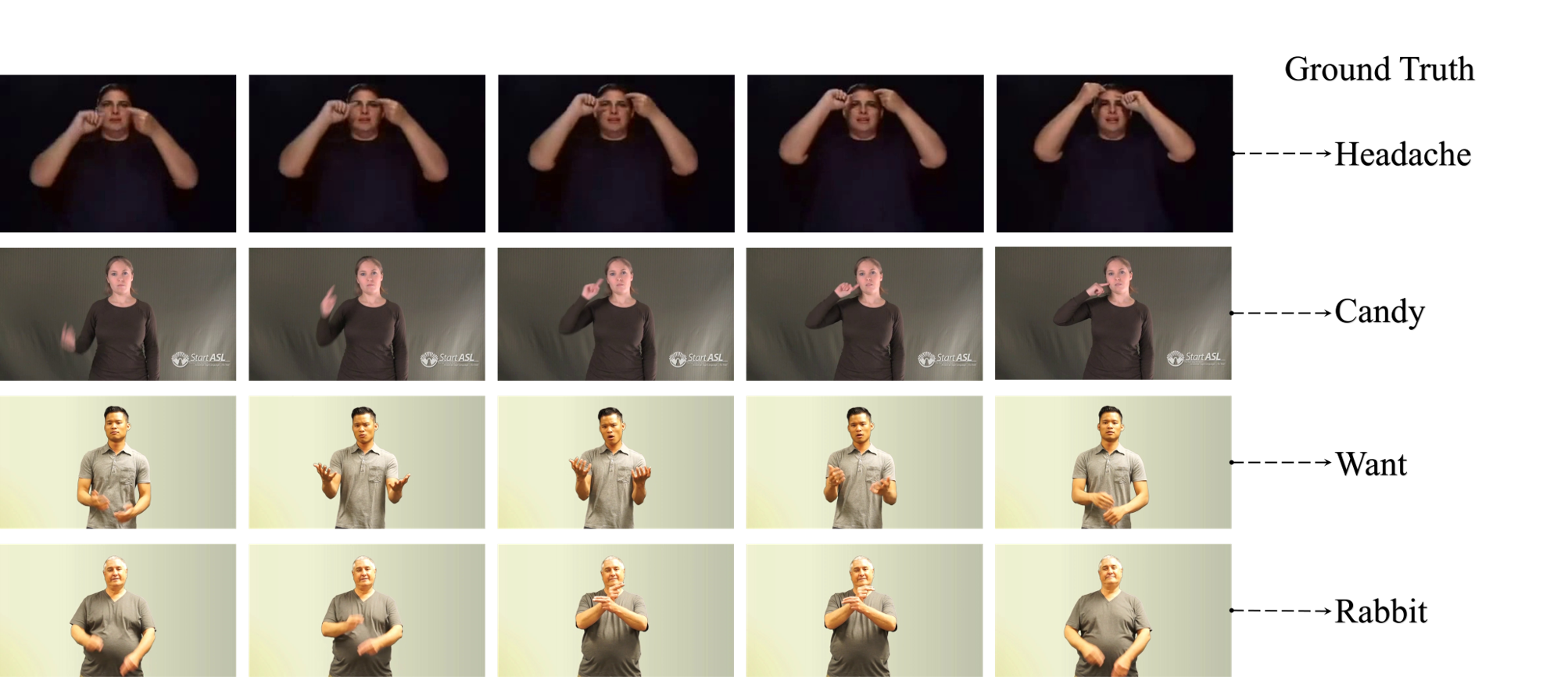}

\end{center}
   \caption{Videos and corresponding ground truth, showing similarities to the predicted glosses in Fig. \ref{fig:res}.}
\label{fig:res_only_gt}
\end{figure*}

\subsection{Implementation details}
The proposed GCN-BERT model has been implemented using PyTorch \cite{NEURIPS2019_9015}. In sign language, different meanings have very similar sign gestures and the difference can only be made out using the contextual information. Hence, following \cite{wlasl}, we use top-\textit{K} accuracy to evaluate the performance of the model. We provide an evaluation of the proposed method using three different values of \textit{K}, specifically {1, 5, 10}. We train the model for 100 epochs using using Adam optimizer with an initial learning rate of $10^{-3}$, a weight decay of $10^{-8}$. Following \cite{wlasl}, we extract 13 upper body keypoints and 21 keypoints corresponding to each hand from each frame of the video using OpenPose \cite{cao}.

\subsection{Results and Analysis}

Table \ref{tab:table_pose_results} shows a comparison of our model with the existing pose-based architectures. The results show that the proposed GCN-BERT model improves the existing state-of-the-art pose-based sign language recognition by a significant margin. This indicates that modelling spatial and temporal relationships separately and explicitly with GCN and BERT respectively improves the prediction accuracy.

\subsection{Qualitative Analysis}

In Figure \ref{fig:res}, we show various videos along with their ground truths and predicted glosses. Though our architecture is pose-based, we show the RGB frames of the video in Figures \ref{fig:res} \& \ref{fig:res_only_gt} for better understanding of the reader. In Figure \ref{fig:res}, we observe that for predicting the word `Doctor', the model is able to capture the temporal dependencies accurately and grasp even the slightest movement in the hand. As we can see from Figures \ref{fig:res} and \ref{fig:res_only_gt}, the signs for the words `halloween' and `headache' are very similar and can be confused even by a human, so it is natural to expect the model to confuse between them (can be seen by observing the top-5 predictions for `halloween' in Figure \ref{fig:res}). We see similar trend for other videos corresponding to `back', `here' and `dark'. For the word `back', we can observe that there is a very subtle difference with the top prediction - `candy'. Given that the signer is also slightly rotated in the frame leads to very similar poses for the words `black' and `candy' making it hard to differentiate. Also, in-plane and out-of-plane movements are not being differentiated by the model due to the fact that we are only utilizing the 2D spatial information. Figure \ref{fig:res} shows a few more signs for which the topmost prediction is not the ground truth. Figure \ref{fig:res_only_gt} contains the videos for the predicted words for comparison of the videos.

In Table \ref{tab:table_pose_results}, we see the effect of increasing the vocabulary size (number of classes) on the performance of the model. Increasing the vocabulary size contributes to a fall in the accuracy. This happens because the dataset consists of ambiguous signs and their meaning depends on the context. Increasing the number of classes, also increases the number of such ambiguous signs, leading to a fall in the accuracy. Based on the observations, we can say that the performance on a smaller dataset does not scale well with a larger dataset. 

\subsection{Conclusion and Future Work}
This work addresses the fundamental problem of sign language recognition in order to bridge the communication barriers between hearing and vocally impaired people, and the rest of the society. Previous works concerned with this problem either jointly considered both spatial and temporal information or relied mainly on the temporal information. To tackle this issue, this paper proposes a novel pose-based architecture for word-level sign language recognition which aims to predict the meaning of the sign language videos. Further, we showed that modelling spatial and temporal information separately with GCN and BERT provides drastic performance gains over the existing state-of-the-art pose-based models. We validated our model on  one of the largest publicly available sign language datasets to show the efficacy of our model. 
% Though our pose-based method improves the accuracy, holistic-image feature based methods provide better performances \cite{wlasl_cvpr} as they include comprehensive features of the image. 
As a part of the future work, we plan to include image-based features into our model to jointly consider both the pose and image related information in order to comprehend the sign language videos.

\section*{Acknowledgement}
We would like to thank our colleagues, Naveen Madapana, Eleonora Giunchiglia, and Aishwarya Chandrasekaran for their constructive feedback.

{\small
\bibliographystyle{ieee_fullname}
\bibliography{egbib}
}

%------------------------------------------------------------------------

\end{document}